\pdfoutput=1

\documentclass[11pt]{article}

\usepackage{naacl2021}
\usepackage{times}
\usepackage{latexsym}

\usepackage[T1]{fontenc}

\usepackage[utf8]{inputenc}
\usepackage{algorithm}
\usepackage{algpseudocode}
\usepackage{algcompatible}
\usepackage{todonotes}
\usepackage{booktabs}

\usepackage{amssymb}
\usepackage{url}
\usepackage{comment}
\usepackage{mathtools,amssymb}
\usepackage{wrapfig,lipsum,booktabs}
\usepackage{hyperref}
\usepackage{multirow}
\usepackage{colortbl}
\usepackage{arydshln}

\usepackage{microtype}
\usepackage{hyperref}

\newcommand{\mz}[1]{}
\newcommand{\gn}[1]{}
\newcommand{\gz}[1]{}

\title{MetaXL: Meta Representation Transformation for Low-resource Cross-lingual Learning}

\author{Mengzhou Xia$^\S$ \thanks{\hspace{1.5mm}Most of the work was done while the first author was an intern at Microsoft Research.} \quad
 Guoqing Zheng$^\ddag$  \quad
 {\bf Subhabrata Mukherjee$^\ddag$ \quad
 Milad Shokouhi$^\ddag$ \quad} \\
 {\bf Graham Neubig$^\dag$ \quad
 Ahmed Hassan Awadallah$^\ddag$}
\\
  $^\S$Princenton University \qquad$^\dag$Carnegie Mellon University \qquad$^\ddag$Microsoft Research\\ 
  mengzhou@princeton.edu, \{zheng, submukhe, milads\}@microsoft.com \\
  gneubig@cs.cmu.edu, hassanam@microsoft.com
}

\begin{document}

\maketitle

\begin{abstract}
The combination of multilingual pre-trained representations and cross-lingual transfer learning is one of the most effective methods for building functional NLP systems for low-resource languages. However, for extremely low-resource languages without large-scale monolingual corpora for pre-training or sufficient annotated data for fine-tuning, transfer learning remains an under-studied and challenging task.  Moreover, recent work shows that multilingual representations are surprisingly disjoint across languages~\citep{singh2019bert}, bringing additional challenges for transfer onto extremely low-resource languages. In this paper, we propose MetaXL, a meta-learning based framework that \textit{learns to transform} representations judiciously from auxiliary languages to a target one and brings their representation spaces closer for effective transfer. Extensive experiments on real-world low-resource languages -- without access to large-scale monolingual corpora or large amounts of labeled data -- for tasks like cross-lingual sentiment analysis and named entity recognition show the effectiveness of our approach. Code for MetaXL is publicly available at \url{{github.com/microsoft/MetaXL}}.
\end{abstract}

\section{Introduction}
\begin{figure}
\centering
\includegraphics[width=0.48\textwidth]{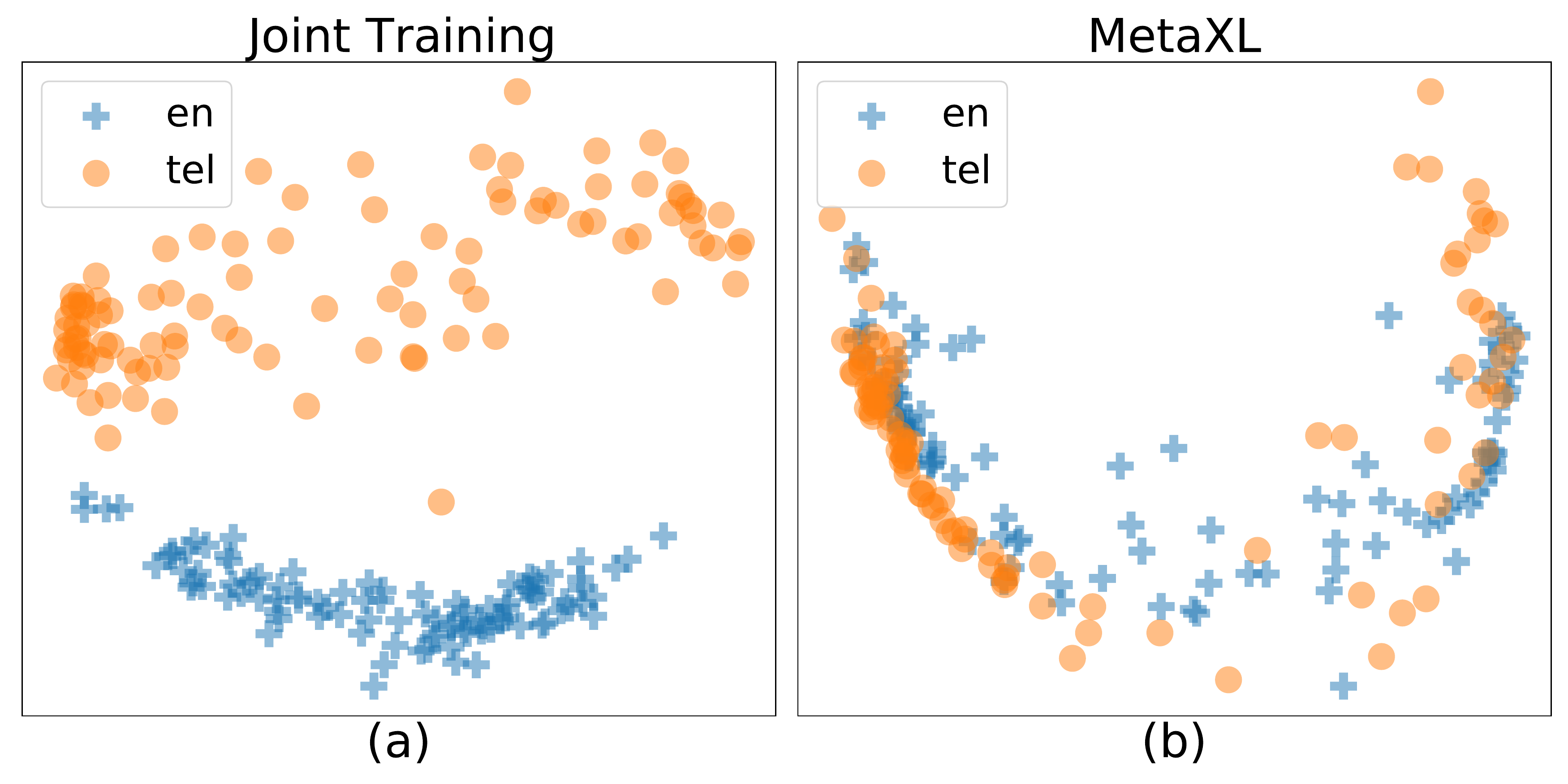}              
\caption{First two principal components of sequence representations (corresponding to \texttt{[CLS]} tokens) of Telugu and English examples from a jointly fine-tuned mBERT and a MetaXL model for the task of sentiment analysis. MetaXL pushes the source (EN) and target (TEL) representations closer to realize a more effective transfer. The Hausdorff distance between the source and target representations drops from $0.57$ to $0.20$ with F1 score improvement from $74.07$ to $78.15$.}
\label{fig:main}
\end{figure}

Recent advances in multilingual pre-trained representations have enabled success on a wide range of natural language processing (NLP) tasks for many languages. However, these techniques may not readily transfer onto extremely low-resource languages, where: (1) large-scale monolingual corpora are not available for pre-training and (2) sufficient labeled data is lacking for effective fine-tuning for downstream tasks. 
For example, multilingual BERT (mBERT)~\citep{devlin2018bert} is pre-trained on $104$ languages with many articles on Wikipedia and XLM-R~\citep{conneau2020unsupervised} is pre-trained on $100$ languages with CommonCrawl Corpora. However, these models still leave behind more than $200$ languages with few articles available in Wikipedia, not to mention the $6,700$ or so languages with no Wikipedia text at all~\cite{artetxe2020call}. Cross-lingual transfer learning for these extremely low-resource languages is essential for better information access but under-studied in practice ~\cite{hirschberg2015advances}. Recent work on cross-lingual transfer learning using pre-trained representations mainly focuses on transferring across languages that are already covered by existing representations~\cite{wu2019beto}.
In contrast, existing work on transferring to languages without significant monolingual resources tends to be more sparse and typically focuses on specific tasks such as language modeling \citep{adams2017cross} or entity linking \citep{zhou2019towards}. 

Building NLP systems in these settings is challenging for several reasons. First, a lack of sufficient annotated data in the target language prevents effective fine-tuning. Second, multilingual pre-trained representations are not directly transferable due to language disparities. Though recent work on cross-lingual transfer mitigates this challenge, it still requires a sizeable monolingual corpus to train token embeddings~\cite{artetxe2019cross}. As noted, these corpora are difficult to obtain for many languages~\cite{artetxe2020call}. 

Additionally, recent work~\cite{singh2019bert} shows that contextualized representations of different languages do not always reside in the same space but are rather partitioned into clusters in multilingual models. This representation gap between languages suggests that joint training with combined multilingual data may lead to sub-optimal transfer across languages. This problem is further exacerbated by the, often large, lexical and syntactic differences between languages with existing pre-trained representations and the extremely low-resource ones. 
Figure \ref{fig:main}(a) provides a visualization of one such example of the disjoint representations of a resource-rich auxiliary language (English) and resource-scarce target language (Telugu). 

We propose a meta-learning based method, MetaXL, to bridge this representation gap and allow for effective cross-lingual transfer to extremely low-resource languages. MetaXL {\em learns to transform} representations from auxiliary languages in a way that maximally facilitates transfer to the target language. Concretely, our meta-learning objective encourages transformations that increase the alignment between the gradients of the source-language set with those of a target-language set. Figure \ref{fig:main}(b) shows that MetaXL successfully brings representations from seemingly distant languages closer for more effective transfer.



We evaluate our method on two tasks: named entity recognition (NER) and sentiment analysis (SA). Extensive experiments on $8$ low-resource languages for NER and $2$ low-resource languages for SA show that MetaXL significantly improves over strong baselines by an average of 2.1 and 1.3 F1 score with XLM-R as the multilingual encoder. 


\section{Meta Representation Transformation}

\begin{figure*}[t]
  \centering
  \includegraphics[width=\textwidth]{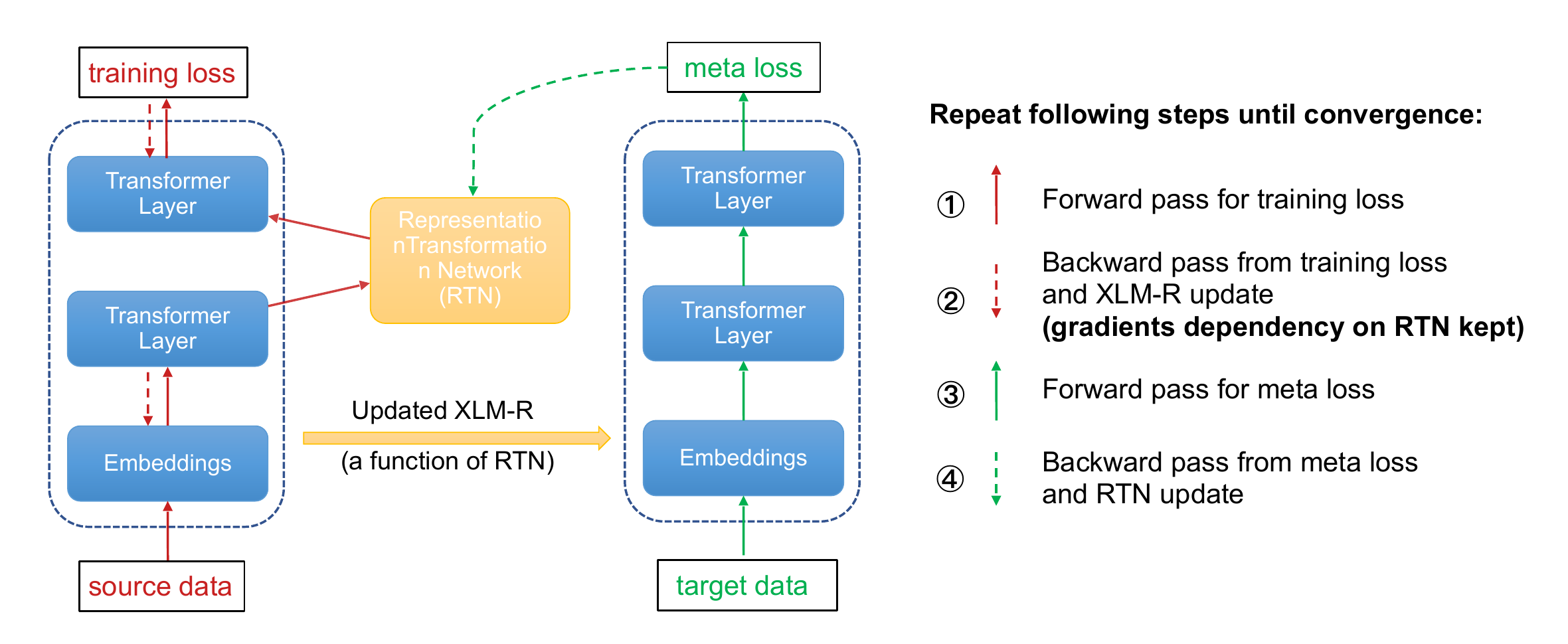}
  \caption{Overview of MetaXL. For illustration, only two Transformer layers are shown for XLM-R, and the representation transformation network is placed after the first Transformer layer. \textcircled{\raisebox{-0.9pt}{1}} source language data passes through the first Transformer layer, through the current representation transformation network, and finally through the remaining layers to compute a training loss with the corresponding source labels. \textcircled{\raisebox{-0.9pt}{2}} The training loss is back-propagated onto all parameters, but only parameters of XLM-R are updated. The updated weights of XLM-R are a function of the current representation transformation network due to gradient dependency (highlighted by the light-purple background of the updated XLM-R). \textcircled{\raisebox{-0.9pt}{3}} A batch of target language data passes through the updated XLM-R and the meta loss is evaluated with the corresponding labels. \textcircled{\raisebox{-0.9pt}{4}} The meta loss is back-propagated into the representation transformation network, since the meta-loss is in effect a function of weights from that network, and only the representation transformation network is updated.} 
  \label{fig:model}
\end{figure*}

\subsection{Background and Problem Definition}



The standard practice in cross-lingual transfer learning is to fine-tune a pre-trained multilingual language model $f_\theta$ parameterized by $\theta$, (e.g. XLM-R and mBERT) with data from one or more auxiliary languages~\footnote{We also refer to auxiliary languages as source languages as opposed to target languages.} and then apply it to the target language. This is widely adopted in the zero-shot transfer setup where no annotated data is available in the target language. The practice is still applicable in the few-shot setting, in which case a small amount of annotated data in the target language is available.

In this work, we focus on cross-lingual transfer for extremely low-resource languages where only a small amount of unlabeled data and task-specific annotated data are available. That includes languages that are not covered by multilingual language models like XLM-R (e.g., Maori or Turkmen), or low-resource languages that are covered but with many orders of magnitude less data for pre-training (e.g., Telegu or Persian). We assume the only target-language resource we have access to is a small amount of task-specific labeled data. 

More formally, given: (1) a limited amount of annotated task data in the target language, denoted as $\mathcal{D}_t = \{(x_t^{(i)}, y_t^{(i)}); i \in [1, N]\}$, (2) a larger amount of annotated data from one or more source language(s), denoted as ${\mathcal{D}_s} = \{(x_s^{(j)}, y_s^{(j)}); j \in [1, M]\}$ where $M \gg N$ and (3) a pre-trained model $f_\theta$, which is not necessarily trained on any monolingual data from the target language -- our goal is to adapt the model to maximize the performance on the target language. 

When some target language labeled data is available for fine-tuning, a standard practice is to jointly fine-tune (JT) the multilingual language model using a concatenation of the labeled data from both the source and target languages $\mathcal{D}_s$ and $\mathcal{D}_t$. The representation gap \cite{singh2019bert} between the source language and target language in a jointly trained model brings additional challenges, which motivates our proposed method. 
 
\subsection{Representation Transformation}
The key idea of our approach is to explicitly learn to transform source language representations, such that when training with these transformed representations, the parameter updates benefit performance on the target language the most. On top of an existing multilingual pre-trained model, we introduce an additional network, which we call the \emph{representation transformation network} to model this transformation explicitly.

The representation transformation network models a function $g_\phi:\mathbb{R}^d\rightarrow\mathbb{R}^d$, where $d$ is the dimension of the representations. Conceptually, any network with proper input and output sizes is feasible. We opt to employ a two-layer feed-forward network, a rather simple architecture with the intention to avoid heavy parameter overhead on top of the pre-trained model. The input to the representation transformation network is representations from any layer of the pre-trained model. By denoting representations from layer $i$ as ${h_i\in\mathbb{R}^d}$, we have a parameterized representation transformation network as follows:
\begin{gather}
    g_\phi (h_i) = w_2^T (\text{ReLU}(w_1^T h_i + b_1)) + b_2
\end{gather}
where $\phi=\{w_1, w_2, b_1, b_2|w_1\in\mathbb{R}^{d\times r}, w_2\in\mathbb{R}^{r\times d}, b_1\in\mathbb{R}^r, b_2\in\mathbb{R}^d\}$ is the set of parameters of the representation transformation network. In practice, we set $r$ to be bottlenecked, i.e.~$r<d$, so the representation transformation network first compresses the input representation and then projects back onto the original dimension of the input representation.

As shown in \autoref{fig:model}, by assuming that the base model has $N$ layers, a source example $(x_s, y_s) \in D_s$ passes through the first $i$ layers, then through the representation transformation network, finally through the last $N-i$ layers of the base model. We denote the final logits of this batch as $f(x_s; \theta, \phi)$, encoded by both the base model and the representation transformation network. In contrast, for a target example $x_t, y_t \in D_t$, we only pass it through the base model as usual, denoted as $f(x_t; \theta)$.

\begin{algorithm*}[t]
\caption{Training procedure for MetaXL}
\label{alg}
\begin{algorithmic}[1]
\Statex{\textbf{Input}: Input data from the target language $D_t$ and the source language $D_s$}
\State{Initialize base model parameters $\theta$ with pretrained XLM-R weights, initialize parameters of the representation transformation network $\phi$ randomly}
\WHILE{not converged}
\State{Sample a source batch $(x_s, y_s)$ from $D_s$ and a target batch $(x_t, y_t)$  from $D_t$;}
\State{Update $\theta$:  $\theta^{(t+1)} = \theta^{(t)} - \alpha \nabla_\theta \mathcal{L}(x_s; \theta^{(t)}, \phi^{(t)})$}
\State{Update $\phi$: $\phi^{(t+1)} = \phi^{(t)} - \beta \nabla_\phi \mathcal{L}(x_t; \theta^{(t)} - \alpha \nabla_\theta \mathcal{L}(x_s; \theta^{(t)}, \phi^{(t)}))$}
\ENDWHILE
\end{algorithmic}
\end{algorithm*}
Ideally, suppose that we have a representation transformation network that could properly transform representations from a source language to the target language. In that case, the source data can be almost equivalently seen as target data on a representation level. Unfortunately, we cannot train such a representation transformation network in a supervised manner without extensive parallel data.

Architecturally, the representation transformation network adopts a similar structure to existing works on language and task adapters for cross-lingual and multi-task transfer~\cite{pfeiffer-etal-2020-adapterhub}, a simple down- and up-projection of input representations. Nevertheless, beyond network architecture, the goal and training procedure of the two approaches are significantly different. Adapters are typically trained to encode task or language-specific information by fixing the rest of the model and updating the parameters of the adapters only. 
Adapters allow training parameter-efficient models that could be flexibly adapted to multiple languages and tasks. While in our proposed method, we use the representation transfer network at training time to adjust the training dynamics to maximally improve test-time performance on the target language. The optimization procedure and the function of the representation transformation network will be discussed in more detail in the next section.

\subsection{Optimization}
The training of the representation transformation network conforms to the following principle: \textit{If the representation transformation network $g_\phi$ effectively transforms the source language representations, such transformed representations $f(x_s; \phi, \theta)$ should be more beneficial to the target task than the original representations $f(x_s; \theta)$, such that the model achieves a smaller evaluation loss $\mathcal{L}_{\mathcal{D}_t}$ on the target language.} This objective can be formulated as a bi-level optimization problem:
\begin{align}
  \label{eq:opt}
  &\min_{\phi}\,\mathcal{L}_{\mathcal{D}_t}\left(f( x_t; \theta^*(\phi)), y_t\right)\\
  \mbox{s.t. }&\,\, \theta^*(\phi) =\arg\min_{\theta}\mathcal{L}_{\mathcal{D}_s}\left(f(x_s; \phi, \theta), y_s\right)\nonumber
\end{align}
where $\mathcal{L}(\cdot)$ is the task loss function.
In this bi-level optimization, the parameters $\phi$ of the representation transformation network are the meta parameters, which are only used at training time and discarded at test time. Exact solutions require solving for the
optimal $\theta^*$ whenever $ \phi$ gets updated. This is computationally infeasible,
particularly when the base model $f$ is complex, such as
a Transformer-based language model. Similar to existing work involving such optimization problems~\cite{finn2017model, liu2018darts, shu2019meta,zheng2021meta}, instead of solving the optimal $\theta^*$ for any given $\phi$, we adopt a one-step stochastic gradient descent update for $\theta$ as an estimate to the optimal base model for a given $\phi$:
\begin{align}
    \theta' = \theta - \alpha \nabla_\theta \mathcal{L}_{\mathcal{D}_s}(f(x_s; \phi, \theta), y_s)
    \label{eq:theta_update}
\end{align}
where $\mathcal{L}_{\mathcal{D}_s}(x_s; )$ is the loss function of the lower problem in \autoref{eq:opt} and $\alpha$ is the corresponding learning rate. Note that the resulting $\theta'$ is in effect a function of $\phi$. We then evaluate the updated weights $\theta'$ on data $x_t$ from the target language for updating $g_\phi$:
\begin{align}
    \phi' = \phi - \beta \nabla_\phi  \mathcal{L}_{\mathcal{D}_t}(f(x_t; \theta'), y_t)
    \label{eq:phi_update}
\end{align}
where $\mathcal{L}_{\mathcal{D}_t}(x_t; \cdot)$ is the loss function of the upper problem in \autoref{eq:opt} and $\beta$ is its corresponding learning rate. Note that the meta-optimization is performed over the parameters of the representation transformation network $g_\phi$ whereas the objective is calculated solely using the updated parameters of the main architecture ${\theta'}$. By plugging \autoref{eq:theta_update} into \autoref{eq:phi_update}, we can further expand the gradient term $\nabla_\phi \mathcal{L} (f(x_t; \theta'), y_t)$. We omit $f$ and $y$ in the following derivative for simplicity.
\begin{gather}
\begin{aligned}
    &\nabla_\phi \mathcal{L}_{\mathcal{D}_t} (x_t; \theta')  \label{eq:meta_update}    \\ 
    = & \nabla_\phi \mathcal{L}_{\mathcal{D}_t} (x_t; \theta - \alpha\nabla_\theta \mathcal{L}_{\mathcal{D}_s}(x_s; \theta, \phi)) \\
     = & - \alpha \nabla^2_{\phi, \theta} \mathcal{L}_{\mathcal{D}_s} (x_s; \theta, \phi)\nabla_\theta \mathcal{L}_{\mathcal{D}_t}(x_t; \theta') \\
     = & - \alpha \nabla_\phi (\nabla_\theta \mathcal{L}_{\mathcal{D}_s}(x_s; \theta, \phi)^T \nabla_\theta \mathcal{L}_{\mathcal{D}_t} (x_t; \theta')) \nonumber
\end{aligned}
\end{gather}
During training, we alternatively update ${\theta}$ with \autoref{eq:theta_update} and $\phi$ with \autoref{eq:phi_update} until convergence. We term our method MetaXL, for its nature to leverage Meta-learning for extremely low-resource cross(X)-Lingual transfer. Both \autoref{fig:model} and Algorithm \autoref{alg} outline the procedure for training MetaXL.


\section{Experiments}
\subsection{Data}

We conduct experiments on two diverse tasks, namely, sequence labeling for Named Entity Recognition (NER) and sentence classification task for Sentiment Analysis (SA). 
For the NER task, we use the cross-lingual Wikiann dataset \cite{pan2017cross}. For the sentiment analysis task, we use the English portion of Multilingual Amazon Reviews Corpus (MARC) \cite{keung-etal-2020-multilingual} as the high-resource language and product review datasets in two low-resource languages, Telugu and Persian \cite{gangula2018resource, hosseini2018sentipers}.

\newcolumntype{R}[1]{>{\RaggedLeft\arraybackslash}p{#1}}
\begin{table}[]
\small
\centering
\begin{tabular}{llll}
\toprule
\multirow{2}{*}{Language}         & \multirow{2}{*}{Code} & Language  & Related \\
& & Family &   Language \\ \midrule
Quechua          & qu   & Quechua  & Spanish       \\
Min Dong & cdo  & Sino-Tibetan    & Chinese\\
Ilocano          & ilo  & Austronesian & Indonesian   \\
Mingrelian       & xmf  & Kartvelian & Georgian      \\
Meadow Mari             & mhr  & Uralic & Russian         \\
Maori            & mi   & Austronesian   & Indonesian \\
Turkmen          & tk   & Turkic   & Turkish       \\
Guarani          & gn   & Tupian & Spanish          \\ \bottomrule
\end{tabular}
\caption{Target language information on the NER task. The data set size of the these languages is 100.}
\label{tab:ner_data}
\end{table}

\begin{table*}[t]
\centering
\begin{tabular}{lllccccccccc}
\toprule
& Source &     Method           & qu    & cdo   & ilo   & xmf   & mhr   & mi    & tk    & gn    & average \\ \midrule
(1)&  - & target               & 57.14 & 37.72 & 61.32 & 59.07 & 55.17 & 76.27 & 55.56 & 48.89 & 56.39 \\ \midrule
\multirow{2}{*}{(2)}&\multirow{2}{*}{ English} & JT & 66.10 & 55.83 & 80.77 & 69.32 & 71.11 & 82.29 & 61.61 & 65.44 & 69.06   \\
& & MetaXL & 68.67 & 55.97 & 77.57 & 73.73 & 68.16 & 88.56 & 66.99 & 69.37 & \textbf{71.13}   \\ \midrule
\multirow{2}{*}{(3)} & \multirow{2}{*}{ Related} & JT  & 79.65 & 53.91 & 78.87 & 79.67 & 66.96 & 87.86 & 64.49 & 70.54 & 72.74   \\
& & MetaXL & 77.06 & 57.26 & 75.93 & 78.37 & 69.33 & 86.46 & 73.15 & 71.96 & \textbf{73.69}  \\
\bottomrule
\end{tabular}
\caption{F1 for NER across three settings where we, (1) only use the target language data; (2) use target language data along with 5k examples of English; (3) use the target language data along with 5k examples of a related language. JT stands for joint training and MetaXL stands for Meta Representation Transformation. We bold the numbers with a better average performance in each setting.}
\label{tab:ner_bilingual}   
\end{table*}


\begin{table}[t]
\centering
\begin{tabular}{llcc} \toprule
  & Method             & tel   & fa    \\ \midrule
(1) & target only  &    86.87 & 82.58 \\ \midrule
\multirow{2}{*}{(2)} & JT & 88.68 & 85.51 \\
& MetaXL & \textbf{89.52} & \textbf{87.14} \\
\bottomrule
\end{tabular}
\caption{F1 for sentiment analysis on two settings using (1) only the target language data; (2) target language data along with 1k examples of English.}
\label{tab:sent_bilingual}
\end{table}

\paragraph{WikiAnn} WikiAnn \cite{pan2017cross} is a multilingual NER dataset constructed with Wikipedia articles and anchor links. We use the train, development and test partitions provided in \citet{rahimi2019massively}. The dataset size ranges from 100 to 20k for different languages.
\paragraph{MARC} The Multilingual Amazon Reviews Corpus \cite{keung-etal-2020-multilingual} is a collection of Amazon product reviews for multilingual text classification. The dataset contains reviews in English, Japanese, German, French, Spanish, and Chinese with five-star ratings. Each language has 200k examples for training. Note that we only use its English dataset.
\paragraph{SentiPers} SentiPers \cite{hosseini2018sentipers} is a sentiment corpus in Persian (fa) consisting of around 26k sentences of users’ opinions for digital products. Each sentence has an assigned quantitative polarity from the set of $\{-2, -1, 0, 1, 2\}$.

\paragraph{Sentiraama} Sentiraama \cite{gangula2018resource} is a sentiment analysis dataset in Telugu (tel), a language widely spoken in India. The dataset contains example reviews in total, labeled as either positive or negative. 

\paragraph{Pre-processing} For SA, we use SentiPers and Sentiraama as target language datasets and MARC as the source language dataset. To unify the label space, we curate MARC by assigning negative labels to reviews rated with 1 or 2 and positive labels to those rated with 4 or 5. We leave out neutral reviews rated with 3.  For SentiPers, we assign negative labels to reviews rated with -1 and -2 and positive labels to those rated with 1 or 2. For SentiPers, though the dataset is relatively large, we mimic the low-resource setting by manually constructing a train, development, and test set with 100, 1000, and 1000 examples through sampling. For Sentiraama, we manually split the dataset into train, development, and test subsets of 100, 103, and 100 examples.\footnote{Details of data splits can be found at \url{github.com/microsoft/MetaXL}.} 

\subsection{Experimental Setup}
\begin{table*}[]
\centering
\begin{tabular}{lccc|lccc|lccc}
\toprule
    & \multicolumn{3}{c|}{NER (average)} &   &   \multicolumn{3}{c|}{SA (tel)}& &  \multicolumn{3}{c}{SA (fa)} \\ 
    \cmidrule(lr){1-4} \cmidrule(lr){5-8} \cmidrule(lr){9-12}
\# en    & JT            & MetaXL   & $\Delta$ &  \# en   & JT       & MetaXL   & $\Delta$ &\# en &  JT       & MetaXL & $\Delta$ \\ \midrule
5k  & 69.06 & 71.13 & +2.07 &  1k  &  88.68 & 90.53 & +1.85 & 1k & 85.51 & 87.14 & +1.63\\
10k & 70.11 & 71.63 & +1.52 & 3k & 87.13 & 87.23 & +0.10 & 3k & 82.88 & 86.19 & +3.31 \\
20k & 72.31 & 73.36 & +1.05 & 5k  & 84.91 & 85.71 & +0.80 & 5k & 86.34 & 85.63 & -0.71  \\ \bottomrule
\end{tabular}
\caption{F1 on various source language transfer data sizes. \# en denotes the number of English examples used for transfer. $\Delta$ denotes the improvement of MetaXL over the joint training baseline. RTN is placed after 12th layer.}
\label{tab:datasize}
\end{table*}


\paragraph{Base Model} We use mBERT\footnote{XLM-R as a base model leads to significantly better results for both baselines and MetaXL than mBERT, thus we mainly present results with XLM-R in the main text. Detailed results on mBERT can be found in \autoref{app:mbert}}~\citep{devlin2018bert} and XLM-R \cite{conneau2020unsupervised} as our base models, known as the state-of-the-art multilingual pre-trained model. However, our method is generally applicable to all types of Transformer-based language models.

\paragraph{Target Language} For NER, we use the same 8 low-resource languages as \citet{pfeiffer-etal-2020-mad}, summarized in \autoref{tab:ner_data}. These languages have only 100 examples in the WikiAnn dataset and are not included for pre-training XLM-R.  For SA, Persian and Telugu are the target languages. For both tasks under any setting, we only use a fixed number of 100 examples for each target language.


\paragraph{Source Language} The selection of source languages is crucial for transfer learning. We experiment with two choices source languages on NER: English and a related language to the target language. The related language was chosen based on LangRank \cite{lin19acl}, a tool for choosing transfer languages for cross-lingual learning. A list of related languages used for each target is shown in \autoref{tab:ner_data}. In absence of training data that fit our related-language criteria for the low-resource target languages in SA, we use only English as the source language.


\paragraph{Tokenization}
For all languages, either pre-trained with XLM-R or not, we use XLM-R's default tokenizer for tokenizing. We tried with the approach where we train subword tokenizers for unseen languages similar to \citet{artetxe2020call} but obtained  worse results than using the XLM-R tokenizer as is, due to the extremely small scale of target language data. We conjecture that the subword vocabulary that XLM-R learns is also beneficial to encode languages on which it is not even pre-trained on. We leave exploring the best tokenization strategy for leveraging pre-trained model on unseen language as future work.





\section{Results and Analysis}
\subsection{Main Results}



\paragraph{NER} We present results of NER in \autoref{tab:ner_bilingual}, where we use 5k examples from English or a related language as source data. When we only use the annotated data of the target language to fine-tune XLM-R (\textit{target}), we observe that the performance varies significantly across languages, ranging from 37.7 to 76.3 F1 score. Jointly fine-tuning XLM-R with target and source data (\textit{JT)} leads to a substantial average gain of around 12.6 F1 score. Using the same amount of data from a related language (instead of English) is more effective, showing an average improvement of 16.3 F1 score over using target data only. Our proposed method, MetaXL, consistently outperforms the joint training baselines, leading to a significant average gain of 2.07 and 0.95 F1 score when paired with English or related languages, respectively. 




\paragraph{SA} We present results on the task of SA in \autoref{tab:sent_bilingual}, where we use 1K examples from English as source language data. We find that auxiliary data from source languages brings less but still significant gains to the joint training baseline (\textit{JT}) over using target language data only (\textit{target only}), as in the NER task. In addition, MetaXL still outperforms joint training by around 0.9 and 1.6 F1 score on Telugu and Persian. These results support our hypothesis that MetaXL can transfer representations from other languages more effectively. That, in turn, contributes to the performance gain on the target task.
\subsection{Source Language Data Size}
To evaluate how MetaXL performs with different sizes of source language data, we perform experiments on varying the size of source data. For NER, we experiment with 5k, 10k, and 20k source examples. For SA, we test on 1k, 3k and 5k \footnote{No significant gains were observed for any of the models when going beyond 5K examples.} source examples. 

As observed from \autoref{tab:datasize}, MetaXL delivers consistent gains as the size of source data increases over the joint training model (except on fa when using 5k auxiliary data).\footnote{Please refer to Appendix \ref{app:datasize} for full results.} However, the marginal gain decreases as the source data size increases on NER. We also note that MetaXL continues to improve even when joint training leads to a minor performance drop for SA.

\subsection{Placement of Representation Transformation Network}

\begin{table}
\centering
\begin{tabular}{lccc}
\toprule
& \multicolumn{1}{c}{NER} & \multicolumn{2}{c}{SA} \\
\cmidrule(lr){2-2} \cmidrule(lr){3-4}
Method & Average & tel & fa \\ \midrule
JT & 69.06 & 88.68 & 85.51\\ \midrule
MetaXL L0 & 70.02 & 89.52 & 85.41 \\
MetaXL L6 & 70.27 & 86.00 & 85.80 \\
MetaXL L12 & \textbf{71.13} & \textbf{90.53} & \textbf{87.14} \\
MetaXL L0,12 & 69.00 & 84.85 & 86.64\\
\bottomrule
\end{tabular}
\caption{F1 when placing the transfer component at different positions on XLM-R. Under this setting, we use 5k English data for NER and 1K English data for SA. L stands for layer.}
\label{tab:re_placement}
\end{table}

Previous works \citep{jawahar2019does, tenney2019bert} have observed that lower and intermediate layers encode surface-level and syntactic information, whereas top layers are more semantically focused. These findings suggest that the placement of the representation transformation network can potentially affect the effectiveness of transfer. To this end, we conducted experiments with representation transformation networks placed at various depths of the Transformer model. 

Specifically, we experiment with placing the representation transformation network after the 0th (embedding layer), 6th and 12th layer (denoted by L0, L6, L12). We also experiment with placing two identical representation transformation networks after both the 0th and 12th layers. 

As observed from \autoref{tab:re_placement}, transformations at the 12th layer are consistently effective, suggesting that transformation at a higher and more abstract level results in better transfer for both tasks.\footnote{Please refer to Appendix \ref{app:replacement} for full results.} Transferring from lower layers leads to fewer gains for SA, coinciding with the fact that SA is more reliant on global semantic information. Transferring at multiple layers does not necessarily lead to higher performance, possibly because it results in increased instability in the bi-level optimization procedure. 


\subsection{Joint Training with Representation Transformation Networks}
\begin{table}[t]
\centering
\begin{tabular}{llccc}
\toprule
& & \multicolumn{1}{c}{NER} & \multicolumn{2}{c}{SA} \\
\cmidrule(lr){3-3} \cmidrule(lr){4-5}
Layer & Method & Average & tel & fa \\ \midrule
-  & JT & 69.06 & 88.68 & 85.51  \\ \midrule
\multirow{2}{*}{L0} & JT w/ RTN & 59.80 & 63.95 & 72.32 \\
& MetaXL & \textbf{70.02} &  \textbf{89.52} & 85.41\\ \midrule 
\multirow{2}{*}{L12} & JT w/ RTN  & 67.18 & 83.75 & 70.40  \\ 
& MetaXL  & \textbf{71.13} & \textbf{90.53} & \textbf{87.14} \\
\bottomrule
\end{tabular}
\caption{F1 when joint training with and without the representation transformation network in XLM-R. In this setting, we use 5k English examples for NER and 1k English examples for SA. NER results are aggregated over 8 target languages. Bold denotes that MetaXL outperforms both JT and JT w/ RTN baselines.}
\label{tab:jointtraining}
\end{table}

There are two major differences between MetaXL and joint training: (1) source language data undergoes transformation via an augmented representation transformation network; (2) we adopt a bi-level optimization procedure to update the base model and the representation transformation network. To verify that the performance gain from MetaXL is not attributed to increased model capacity, we conduct experiments on joint training using the representation transformation network. Specifically, the forward pass remains the same as MetaXL, whereas the backward optimization employs the standard stochastic gradient descent algorithm. We conduct experiments on placing the representation transformation network after the 0th layer or 12th layer and present results in \autoref{tab:jointtraining} \footnote{Please refer to Appendix \ref{app:jt} for full results.}. 

Interestingly, joint training with the representation transformation network deteriorates the model performance compared to vanilla joint training. Transferring after the 0th layer is even more detrimental than the 12th layer. This finding shows that Transformer models are rather delicate to subtle architectural changes. In contrast, MetaXL breaks the restriction, pushing the performance higher for both layer settings. 

\subsection{Analysis of Transformed Representations}
\begin{figure}[t]
\centering
\includegraphics[width=0.49\textwidth]{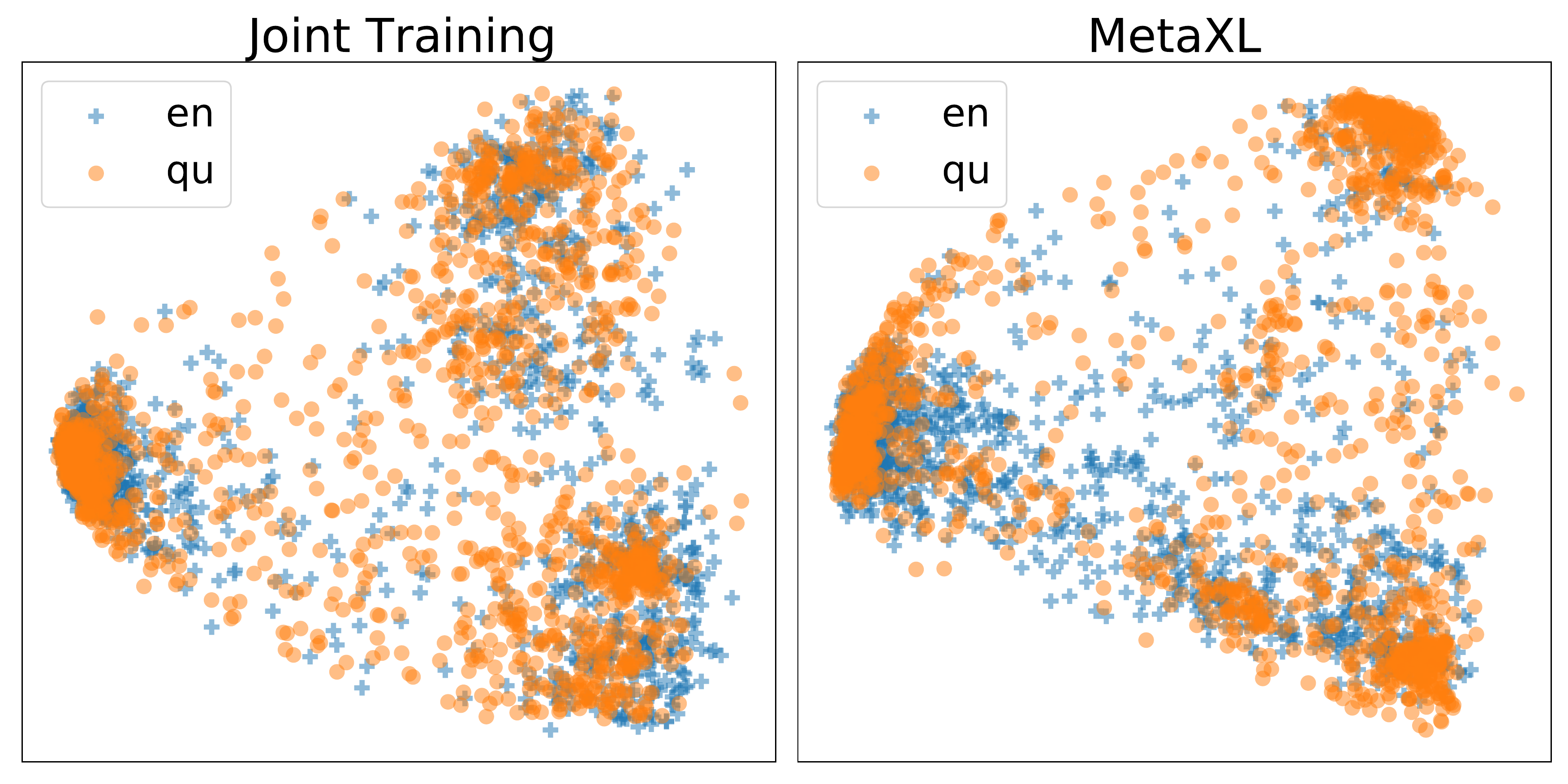}              
\caption{PCA visualization of token-level representations of Quechua and English from the joint training mBERT model on NER. With MetaXL, the Hausdorff distance drops from 0.60 to 0.53 and the F1 score increases from 60.25 to 63.76.}
\label{fig:qu}
\end{figure}

To verify that MetaXL does bring the source and target language spaces closer, we qualitatively and quantitatively demonstrate the representation shift with transformation. In particular, we collect representations of both the source and target languages from the joint training and the MetaXL models, with mBERT as the multilingual encoder, and present the 2-component PCA visualization in \autoref{fig:main} for SA and \autoref{fig:qu} for NER. SA models are trained on Telugu paired with 5k English examples, and NER models are trained on Quechua paired with 5k English. From the figures, MetaXL merges the representations from two languages for SA, but the phenomenon is not as evident for NER.

\citet{singh2019bert} quantitatively analyze mBERT representations with  canonical correlation analysis (CCA).
However, CCA does not suit our case as we do not have access to semantically aligned data for various languages. Thus we adopt Hausdorff distance as a metric that has been widely used in vision and NLP tasks \cite{huttenlocher1993comparing, dubuisson1994modified,patra-etal-2019-bilingual} to measure the distance between two distinct datasets. Informally, the Hausdorff distance measures the average proximity of data representations in the source language to the nearest ones in the target language, and vice versa. Given a set of representations of the source language $\mathcal{S}=\{s_1, s_2, \dots, s_m\}$ and a set of representations of the target language $\mathcal{T}=\{t_1, t_2, \dots, t_n\}$, we compute the Hausdorff distance as follows: 
\begin{align}
     \max\{ \max_{s \in \mathcal{S}}\min_{t \in \mathcal{T}}d(s, t), \max_{t \in \mathcal{T}}\min_{s \in \mathcal{S}}d(s, t)\}
\end{align}
where cosine distance is used as as the inner distance, i.e.,
\begin{align}
d(s,t)\triangleq 1-\cos(s,t)
\end{align}

For SA, we observe a drastic drop of Hausdorff distance from 0.57 to 0.20 and a substantial performance improvement of around 4 F1 score. For NER, we observe a minor decline of Hausdorff distance from 0.60 to 0.53 as the representations are obtained at the token level, leading to a significant performance gain of 3 F1 score. For NER, we observe a correlation of 0.4 between performance improvement and the reduction in representation distance. Both qualitative visualization and quantitative metrics confirm our hypothesis that MetaXL performs more effective transfer by bringing the representations from different languages closer.

\subsection{Additional Results on High-resource Languages}
\begin{table}[h]
\centering
\begin{tabular}{lccccc}
\toprule 
& fr & es & ru & zh  \\ \midrule
JT & 76.50 & 72.87 & 71.14 & 60.62 \\ 
MetaXL & 72.43 & 70.38 & 71.08 & 58.81 \\ 
\bottomrule
\end{tabular}
\caption{F1 on mBERT rich languages in a simulated low-resource setting. }
\label{tab:highresource}
\end{table}

Despite our experiments so far on extremely low-resource languages, given by few labeled data for fine-tuning and limited or no unlabeled data for pre-training, MetaXL is generally applicable to all languages. To better understand the scope of applying MetaXL to languages with varying resources, we perform experiments on 4 target languages that do not belong to our extremely low-resource category for the NER task, namely, Spanish (es), French (fr), Italian (it), Russian (ru) and Chinese (zh). These languages are typically considered high-resource with 20k labeled examples in the WikiAnn datasets and large amount of unlabeled data consumed by mBERT for pre-training. We use only 100 examples for all target languages to mimic the low-resource setting and use 5k English examples for transfer.

As shown in \autoref{tab:highresource}, we found slight performance drop using MetaXL for these high-resource languages. We conjecture that these languages have been learned quite well with the mBERT model during the pre-training phase, therefore, leaving less scope for effective representation transformation in the low-resource setup. Nonetheless, this can be remedied with a back-off strategy by further fine-tuning the resulting model from MetaXL on the concatenated data from both source and target languages to match the performance of JT training. As high-resource languages are out of the scope of this paper, we leave further analysis and understanding of these scenarios for future work.

\section{Related Work}
\label{sec:relatedwork}





\paragraph{Unifying Language Spaces} MetaXL in essence brings the source and target representations closer. Previous works have shown that learning invariant representations across languages leads to better transfer. On the representation level, adversarial training is widely adopted to filter away language-related information \cite{xie2017controllable, chen2018adversarial}. One the form level, \citet{xia2019generalized} show that replacing words in a source language with the correspondence in the target language brings significant gains in low-resource machine translation. 

\paragraph{Adapters}  Adapter networks are designed to encode task \cite{houlsby2019parameter, stickland2019bert, pfeiffer2020adapterfusion}, domain \cite{bapna-firat-2019-simple} and language \cite{pfeiffer-etal-2020-mad} specific information to efficiently share parameters across settings. Though RTN in MetaXL is similar to adapter networks in architecture, in contrast to adapter networks, it plays a more explicit role in transforming representations across languages to bridge the representation gap. More importantly, MetaXL trains the representation transformation network in a meta-learning based paradigm, significantly different from how adapters are trained. 

\paragraph{Meta Learning} MetaXL falls into the category of meta learning for its goal to \textit{learn to transform} under the guidance of the target task. Related techniques have been used in \citet{finn2017model}, which aims to learn a good initialization that generalizes well to multiple tasks and is further extended to low-resource machine translation \cite{gu-etal-2018-meta} and low-resource natural language understanding tasks \cite{dou-etal-2019-investigating}. The bi-level optimization procedure is widely adopted spanning across neural architecture search \cite{liu2018darts}, instance re-weighting~\cite{ren2018learning,shu2019meta}, learning from pseudo labels \cite{pham2020meta} and mitigating negative inference in multilingual systems \cite{wang-etal-2020-negative}. MetaXL is the first to meta learn a network that explicitly transforms representations for cross-lingual transfer on extremely low-resource languages. 

\section{Conclusions and Future Work}

In this paper, we study cross-lingual transfer learning for extremely
low-resource languages without large-scale monolingual corpora for
pre-training or sufficient annotated data for fine-tuning. To allow
for effective transfer from resource-rich source languages and
mitigate the representation gap between multilingual pre-trained
representations, we propose MetaXL to \textit{learn to transform}
representations from source languages that best benefits a given task
on the target language. Empirical evaluations on cross-lingual
sentiment analysis and named entity recognition tasks demonstrate the
effectiveness of our approach. Further analysis on the learned
transformations verify that MetaXL indeed brings the representations
of both source and target languages closer, thereby, explaining the
performance gains. For future work, exploring transfer from multiple
source languages to further improve the performance and investigating the placement of multiple representation transformation networks on multiple
layers of the pre-trained models are both interesting directions to
pursue.

\section*{Acknowledgements}

We thank the anonymous reviewers for their constructive feedback, and Wei Wang for valuable discussions.

\section*{Ethical Considerations}

This work addresses cross-lingual transfer learning onto extremely low-resource languages, which is a less studied area in NLP community. We expect that progress and findings presented in this paper could advocate awareness of advancing NLP for extremely low-resource languages and help improve information access for such under-represented language communities.

The proposed method is somewhat compute-intensive as it requires approximating second-order gradients for updating the meta-parameters. This might impose negative impact on carbon footprint from training the described models. Future work on developing more efficient meta-learning optimization methods and accelerating meta-learning training procedure might help alleviate this concern.

\bibliography{anthology,custom,ref}
\bibliographystyle{acl_natbib}

\clearpage
\appendix
\section{Hyper-parameters}
\label{sec:appendix:hyp}
We use a maximum sequence length of 200 and 256 for NER and AS respectively. We use a bottleneck dimension of $r=384$ and $r=192$  for the representation transformation network, same as \citet{pfeiffer-etal-2020-mad}. During the bi-level optimization process, we adopt a learning rate of 3e-05 for training the main architecture and tuned the learning rate on 3e-5, 1e-6 and 1e-7 for training the representation transformation network. We use a batch size of 16 for NER and 12 for AS, and train 20 epochs for each experiment on both tasks. We use a single NVIDIA Tesla V100 with a 32G memory size for each experiment. For each language, we pick the best model according to the validation performance after each epoch.

\section{Detailed Results on Each Language}

\subsection{Source Data Size}
The full results of using 10k and 20k English examples as transfer data are presented in \autoref{tab:app_datasize}.

\subsection{Placement of RTN}
\label{app:replacement}
The full results of placing the representation transformation network at different layers are presented in \autoref{tab:app_layer}.

\subsection{Joint Training w/ RTN}
\label{app:jt}
The full results of joint training with the representation transformation network are presented in \autoref{tab:jt}.

\section{Additional Results on mBERT}
\label{app:mbert}

\begin{table}[b]
\centering
\begin{tabular}{llcc} \toprule
  & Method             & tel   & fa    \\ \midrule
(1) & target only  &   75.00 & 73.86 \\ \midrule
\multirow{2}{*}{(2)} & JT & 75.13 & 74.81 \\
& MetaXL & \textbf{77.36} & \textbf{76.69} \\
\bottomrule
\end{tabular}
\caption{F1 for sentiment analysis on mBERT on two settings using (1) only the target language data; (2) target language data along with 10k examples of English.}
\label{tab:sa-mbert}
\end{table}

\label{app:datasize}
\begin{table*}[t]
\centering
\begin{tabular}{lllccccccccc}
\toprule
& Source &     Method           & qu    & cdo   & ilo   & xmf   & mhr   & mi    & tk    & gn    & average \\ \midrule
(1) &  - & target only               & 57.14 & 37.72 & 61.32 & 59.07 & 55.17 & 76.27 & 55.56 & 48.89 & 56.39 \\ \midrule
\multirow{2}{*}{(2)} & \multirow{2}{*}{ 10k en} & JT & 71.49 & 50.21 & 76.19 & 73.39 & 66.36 & 89.34 & 66.04 & 67.89 & 70.11   \\
&& MetaXL & 72.57 & 57.02 & 81.55 & 65.56 & 70.18 & 90.64 & 66.98 & 68.54 & \textbf{71.63}   \\ \midrule
\multirow{2}{*}{(3)} & \multirow{2}{*}{ 20k en} & JT  & 73.19 & 53.93 & 88.78 & 71.49 & 62.56 & 90.80 & 68.29 & 69.44 & 72.31   \\
&& MetaXL & 73.04 & 55.17 & 85.99 & 73.09 & 70.97 & 89.21 & 66.02 & 73.39 & \textbf{73.36}  \\
\bottomrule
\end{tabular}
\caption{Experiment results for NER on XLM-R across three settings where we, (1) only use the target language data; (2) use target language data along with 10k examples of English; (3) use target language data along with 20k examples of English. JT stands for joint training}
\label{tab:app_datasize}   
\end{table*}

\begin{table*}[t]
\centering
\begin{tabular}{llccccccccc}
\toprule
Layer &     Method           & qu    & cdo   & ilo   & xmf   & mhr   & mi    & tk    & gn    & average \\ \midrule
- & JT & 66.1 & 55.83 & 80.77 & 69.32 & 71.11 & 82.29 & 61.61 & 65.44 & 69.06 \\ \midrule
L0 &  MetaXL & 70.43 & 54.76 & 77.14 & 66.09 & 68.72 & 89.53 & 63.59 & 69.86 & 70.02   \\
L6 & MetaXL & 65.53 & 56.67 & 78.5 & 72.0 & 68.75 & 88.05 & 65.73 & 66.96 & 70.27   \\ 
L0,12 & MetaXL & 69.83 & 53.97 & 69.44 & 69.26 & 66.96 & 89.41 & 67.92 & 65.18 & 69.00   \\
\bottomrule
\end{tabular}
\caption{Experiment results for NER on XLM-R with RTN placed across multiple layer settings. (All with 5k English examples)}

\label{tab:app_layer}   
\end{table*}

\begin{table*}[t]
\centering
\begin{tabular}{llccccccccc}
\toprule
Layer &     Method           & qu    & cdo   & ilo   & xmf   & mhr   & mi    & tk    & gn    & average \\ \midrule
- & JT & 66.10 & 55.83 & 80.77 & 69.32 & 71.11 & 82.29 & 61.61 & 65.44 & 69.06 \\ \midrule
L0 &  JT w/ RTN & 50.81 & 45.67 & 60.09 & 58.91 & 63.83 & 81.71 & 65.37 & 52.02 & 59.80   \\
L12 & JT w/ RTN & 64.41 & 50.2 & 73.83 & 63.87 & 68.7 & 85.88 & 71.92 & 58.6 & 67.18 \\
\bottomrule
\end{tabular}
\caption{Experiment results for NER on XLM-R, Joint Training (JT) with RTN. (All with 5k English examples)}
\label{tab:jt}   
\end{table*}

We conduct experiments on mBERT \cite{devlin-etal-2019-bert}, which covers 104 languages with most Wikipedia articles. For a language that is not pre-trained with mBERT, we train its subword tokenizer with the task data. Further, we combine the vocabulary from the newly trained tokenizer with the original mBERT vocabulary. A similar approach has been adopted in \cite{artetxe2020call}. \autoref{tab:ner-mbert} and \autoref{tab:sa-mbert} present results for NER and SA respectively where we finetune the tasks on mBERT. Note that the languages of SA are both covered by mBERT and XLM-R, while the languages of NER are not. \autoref{tab:mbertdatasize} show MetaXL results on mBERT with various sizes of source data.
 
Nevertheless, our method consistently brings gains on both tasks. We observe an average of 2 F1 points improvement on NER and 2.0 F1 points improvement on SA. It shows that the improvement brought by our method is consistent across different language models.

\begin{table*}[t]
\centering
\begin{tabular}{lllccccccccc}
\toprule
& Source &     Method           & qu    & cdo   & ilo   & xmf   & mhr   & mi    & tk    & gn    & average \\ \midrule
(1)&  - & target & 58.44 & 26.77 & 63.39 & 32.06 & 53.66 & 82.90 & 52.53 &  46.01 & 51.97 \\ \midrule
\multirow{2}{*}{(2)}&\multirow{2}{*}{ English} & JT & 60.25 & 35.29 & 73.06 & 43.45 & 60.17 & 86.29 & 60.09 & 57.80  & 59.55   \\
& & MetaXL & 63.76 & 38.63 & 76.36 & 45.14 & 60.63 & 88.96 & 64.81 & 54.13 & \textbf{61.55}   \\ 
\bottomrule
\end{tabular}
\caption{NER results on mBERT where we use 5k English examples as auxiliary data and place RTN after 12th layer. }
\label{tab:ner-mbert}   
\end{table*}

\begin{table*}[]
\centering
\begin{tabular}{lccc|lccc|lccc}
\toprule
    & \multicolumn{3}{c|}{NER (average)} &   &   \multicolumn{3}{c|}{SA (tel)}& &  \multicolumn{3}{c}{SA (fa)} \\ 
    \cmidrule(lr){1-4} \cmidrule(lr){5-8} \cmidrule(lr){9-12}
\# en    & JT            & MetaXL   & $\Delta$ &  \# en   & JT       & MetaXL   & $\Delta$ &\# en &  JT       & MetaXL & $\Delta$ \\ \midrule
5k  & 59.55         & 61.55 & +2.00                 & 100 & 75.12    & 77.36 &       +2.24    & 100            & 74.25    & 75.78    &      +1.53                 \\
10k & 62.36         & 63.66 & +1.30                 & 1k  & 74.76    & 76.39 &       +1.63  & 1k         & 74.71    &   75.58  &    +0.87                   \\
20k & 62.39         & 63.38 & +0.99                 & 5k & 74.07    & 78.15 &       +4.08    & 5k            & 74.81    &   76.69  & +1.88                       \\ \bottomrule
\end{tabular}
\caption{F1 on various source language transfer data sizes on mBERT. \# en denotes the number of English examples used for transfer. $\Delta$ denotes the improvement of MetaXL over the joint training baseline.}
\label{tab:mbertdatasize}
\end{table*}

\end{document}